\documentclass{article}
\usepackage{spconf,amsmath,graphicx, booktabs}
\usepackage[super]{nth}

\title{Are cascade dialogue state tracking models speaking out of turn in spoken dialogues?}
\name{Lucas Druart*\textsuperscript{1,2} \quad Léo Jacqmin*\textsuperscript{1,3} \quad Benoit Favre\textsuperscript{3} \quad Lina Maria Rojas-Barahona\textsuperscript{1} \quad Valentin Vielzeuf \textsuperscript{1}}
\address{\textsuperscript{1} Orange Innovation, France \quad
    \textsuperscript{2} LIA - Avignon Université, France \quad \textsuperscript{3} LIS - Aix-Marseille Université, France\\ 
    \small \textsuperscript{1}\texttt{\{lucas1.druart,leo.jacqmin,linamaria.rojasbarahona,valentin.vielzeuf\}@orange.com}\\ \small\textsuperscript{2}\texttt{\{first.last\}@univ-avignon.fr} \quad \small\textsuperscript{3}\texttt{\{first.last\}@lislab.fr} \quad \small*\texttt{equal contribution}}

\begin{document}

    \maketitle
    \begin{abstract}
    In Task-Oriented Dialogue (TOD) systems, correctly updating the system's understanding of the user's needs is key to a smooth interaction. Traditionally TOD systems are composed of several modules that interact with one another. While each of these components is the focus of active research communities, their behavior in interaction can be overlooked. This paper proposes a comprehensive analysis of the errors of state of the art systems in complex settings such as Dialogue State Tracking which highly depends on the dialogue context. Based on spoken MultiWoz, we identify that errors on non-categorical slots' values are essential to address in order to bridge the gap between spoken and chat-based dialogue systems. We explore potential solutions to improve transcriptions and help dialogue state tracking generative models correct such errors.
    
    \end{abstract}
    
    \section{Introduction}
    Task-Oriented Dialogue (TOD) systems are designed to interactively assist a user with a challenging task such as making a reservation at a restaurant or booking a room in a hotel. A typical approach to implement them is to use a modular architecture \cite{gaoNeuralApproachesConversational2018a}. A core component of such systems is Dialogue State Tracking (DST) whose goal is to update a condensed representation of the user needs up to a given turn. Based on this representation, the dialogue system decides which action to take next towards completing the task.
    
    While spoken dialogue was the focus of earlier work \cite{williamsDialogStateTracking2016}, recent work on DST has focused on text inputs with no regard for the specificities of spoken language \cite{fengSequencetoSequenceApproachDialogue2021}. The underlying assumption is that Automatic Speech Recognition (ASR) and DST can be treated separately and advances from both lines of work effectively integrated. However, this approach fails to account for the differences between written and spoken language \cite{faruquiRevisitingBoundaryASR2022}. By studying both ASR and DST jointly, this paper identifies remaining challenges to bridge the gap between spoken and written DST.

    \begin{figure}[t!]
    \centering
    \includegraphics[width=1\linewidth]{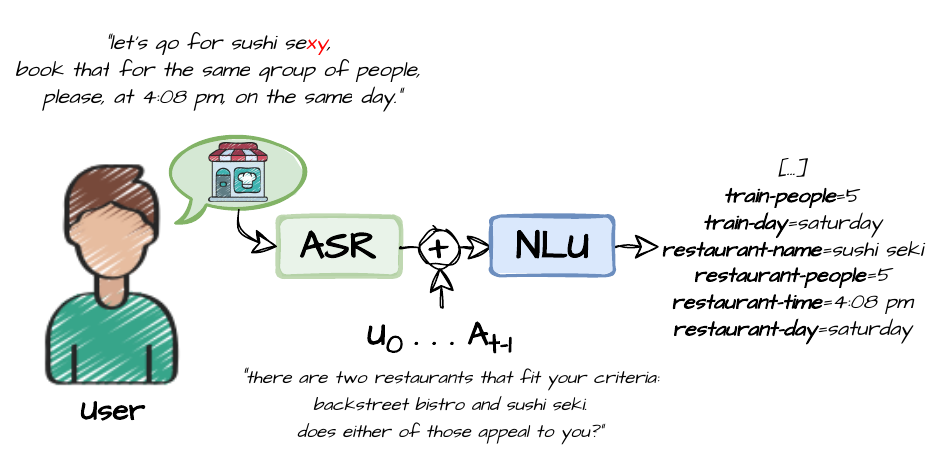}
    \caption{Spoken DST cascade system at inference for turn $t$. Red characters indicate transcription errors. Such examples illustrate the challenge of contextually aware predictions.}
    \label{fig:cascade}
    \end{figure}
    
    DST models are robust to some transcription errors \cite{cho-etal-2022-know}, however we do not yet have a clear classification of the errors and whether they will impact the final output. This paper provides a comprehensive analysis and categorization of errors which occur between the ASR and DST modules. When comparing spoken cascade DST to text DST, we find that the main performance degradation occurs on non-categorical slots which values come from an open set. Though some enhancements to the system such as specific post-processing of the ASR transcriptions and augmenting the DST training data can mitigate some of this degradation, the gap between spoken and text DST remains. Our analysis highlights potential directions to address these issues.
    
    \section{Related Work}

    Early editions of the Dialogue System Technology Challenge (DSTC) focused on spoken dialogue \cite{williamsDialogStateTracking2013,henderson-etal-2014-second}. However, unlike modern DST systems which operate directly on the user utterances \cite{fengSequencetoSequenceApproachDialogue2021}, the proposed DST systems operated on turn-level Spoken Language Understanding (SLU) outputs: at the time the challenge was to account for the uncertainty of the ASR and SLU \cite{hendersonDiscriminativeSpokenLanguage2012,tur2013semantic}, in part due to their lower performances. 
    
    Following this series of work, DST has been mainly studied in the context of chat corpora \cite{wuTransferableMultiDomainState2019,heckTripPyTripleCopy2020}. The more recent editions of DSTC have redirected the focus towards spoken dialogue. DSTC10 \cite{Kim2021HowRR} provided ASR transcriptions on which to build DST adaptation methods for this specific set of transcriptions. The latest edition DSTC11 goes further and provides audio recordings \cite{Soltau2022SpeechAD}. Our analysis is performed in this novel context and proposes a comprehensive study of the sources of errors.

    Both End-to-End (E2E) \cite{E2ESLU} and cascade approaches \cite{CascadeSLU} have been studied extensively to solve different SLU tasks \cite{QinSLUSurvey} with independent utterances. Error analysis \cite{SLU-Impact} of E2E and cascade models using a same Wav2Vec2.0 speech encoder underlines that the E2E fine-tuning degrades the model's transcription capabilities compared to its cascade counterpart, especially on unseen slot-value pairs. 

    This paper focuses on the behavior of SOTA cascade systems (e.g. native ASR correction, coping with hallucinations, learning with synthetic data augmentation) for contextually aware semantic extraction in which the previous dialogue turns are mandatory to correctly process the current one (\textit{e.g.} cross-turn references). 
    
    
    
    \section{Spoken Dialogue State Tracking}
     
    \subsection{Spoken version of MultiWOZ}

    MultiWoz is a human-human chat-based English TOD dataset commonly used for training and evaluating dialogue systems \cite{Budzianowski2018MultiWOZA}. A spoken version with vocalized user turns was published in the context of the \textit{"Speech Aware"} track of DSTC11\footnote{https://dstc11.dstc.community/} \cite{Soltau2022SpeechAD}.
    The user utterances in the dev and test sets (\textbf{Dev}$|$\textbf{Test}) are available as both synthetic and human speech (\textbf{TTS}$|$\textbf{Human}), whereas the training set is only available as synthetic speech. The dataset contains close to 10,000 dialogues with a 80/10/10 train-dev-test split and an average of 13.3 turns per dialogue. 
    Among the pre-defined slots, we can distinguish three groups: categorical slots with a closed set of values, non-categorical slots with an open set of values ($\sim$30\%) and time slots ($\sim$10\%). 
    
    Additionally, due to a high overlap in the value distributions of the training and evaluation sets \cite{qianAnnotationInconsistencyEntity2021} which favors memorization, the values for non-categorical slots in the dev and test sets of the spoken version were replaced with new values from an unknown database. Time slots were also offset by a constant amount. 
    

    Our analysis draws on a turn-level exact match metric known as Joint-Goal Accuracy (JGA$\uparrow$), a Slot Type Accuracy (STA$\uparrow$) and a per Slot-type value Precision (SP$\uparrow$) to distinguish errors.  
    
    \subsection{Automatic Speech Recognition}
    
    We benchmark three different established ASR
    approaches (Robust Wav2Vec \cite{Hsu2021RobustW2}, Conformer-Transducer \cite{conformerTransducer} and OpenAI Whisper \cite{Radford2022RobustSR}) on the Dev-Human data with a common DST model (Table \ref{tab:ASRBench}) and retain the best one for our experiments.
    
    \begin{table}[hbt]
    \centering
    \resizebox{\linewidth}{!}{
    \begin{tabular}{lc|c}
            & \textbf{WER$\downarrow$} & \textbf{JGA$\uparrow$} \\ \midrule
            \textbf{Conformer-Transducer} (600 M) & 10.00 & 27.10 \\  
            \textbf{Robust Wav2Vec} (315 M) & 16.41 & 23.90 \\
            \textbf{Whisper} (769 M) & \textbf{8.04} & \textbf{29.30} \\ \bottomrule
        \end{tabular}
        }
    \caption{WER and JGA of 3 different ASR approaches.}
    \label{tab:ASRBench}
    \end{table}
    

    
    \subsection{Dialogue State Tracking}
    
    The used DST approach is ranked as the best system of the challenge \cite{olisia}. It is more robust to ASR noise compared to extractive approaches \cite{cho-etal-2022-know}. It relies on a pre-trained T5 \cite{RaffelT5} as backbone model which is further fine-tuned on the MultiWOZ dataset.  
    
    The encoder is inputted the entire dialogue history at turn $t$ composed of user and agent turn pairs denoted as $X_t = [U_1, A_1, ..., U_t, A_t]$\footnote{If the entire dialogue history exceeds the model's maximum input length, the input is truncated to include only the most recent turns.}. To preserve the dialogue structure, agent and user turns are separated by the special tokens \texttt{agent:} and \texttt{user:}. The target dialogue state $DS_t$ is linearized as a coma separated list of \texttt{<slot,value>} pairs: \texttt{slot$_1$=value$_1$;...;slot$_n$=value$_n$}.
    
    For ease of comparison, while keeping a low computational cost, all the experiments are conducted with fine-tuned T5-small models on different versions of the training data with replaced values, with a learning rate of $5e^{-4}$ and a batch size of $16$. 
    
    \subsection{Enhancements}
    
    We propose enhancements to this core "vanilla" model and analyze their impact on the different error sources.
    
    \subsubsection{Improving transcriptions}
    
    We infer the original MultiWoz data normalisation rules and apply them to the ASR transcriptions for the DST to be able to rely on the same cues. We use the whisper normalizer and format all time mentions to \texttt{[hours]:[minutes] [am|pm]}. 
    
    Then we propose to correct non-categorical slot values. We first identify the lists of proper nouns in agent and user turns with a BERT based Named-Entity Recognition (NER) model from the NeMo toolkit \cite{NemoToolkit}. We then score each pair of user and agent identified named entities with Character Error Rate (CER) and tune a threshold to replace the user turns' misspelled proper nouns with the corresponding match from the agent turns.
    
    \subsubsection{Improving state tracking}
    \label{sec:dst_improvements}
    
    In order to improve the transcription error robustness of the DST model, we augment the training data with examples which present such errors.
    
    Our first approach consists in simulating ASR errors by injecting three types of character-level errors in the ground-truth transcriptions. We randomly insert, delete, or replace a character in the span corresponding to a non-categorical slot value.
    All three injections can be applied multiple times and combined. New characters are sampled according to a character-level error matrix between Whisper tiny's predicted transcription and the ground-truth ones over the training set. We refer to the DST model fine-tuned on this data as \textbf{Rule-error DST}.
    
    Our second approach (\textbf{TTS-ASR DST}) consists in a TTS-ASR pipeline to simulate more realistic speech transcriptions errors. We synthesize user turns with a Tacotron2-based \cite{shen2018natural} TTS system and transcribe them with an ASR system composed of a CRDNN with CTC/Attention and an RNN LM trained on LibriSpeech. 
    
    
    
    
    
    
    
    \section{Error Analysis}
    
    We present the main results of an oracle (predictions made on ground-truth transcripts), baseline, and enhanced models in Table \ref{tab:main_results}, along with our challenge performance (obtained with larger models) as a frame of reference.
    
    \begin{table}[hbt!]
    \resizebox{\linewidth}{!}{
        \begin{tabular}{lcc|cc}
            & \multicolumn{2}{c}{\textbf{Dev}} & \multicolumn{2}{c}{\textbf{Test}} \\ \midrule
            \textbf{Text Oracle} (JGA/STA) & \multicolumn{2}{c}{50.4 / 65.7} & \multicolumn{2}{c}{47.5 / 61.2} \\
            & \textbf{TTS} & \textbf{Human} & \textbf{TTS} & \textbf{Human} \\ \midrule
            DSTC11 Baseline & \_ & \_ & 27.7 / \_ & 23.6 / \_  \\
            DSTC11 1$^{st}$ place & 47.2 / \_ & 43.2 / \_ & 44.0 / \_ & 39.5 / \_  \\   \midrule
            \textbf{W.+Text DST} & 33.7 / \textbf{65.0} & 27.7 / 63.2  & 31.4 / 61.2 & 26.5 / 58.3    \\
            + normalization & 36.6 / 64.5 & 32.0 / 62.6 & 35.0 / 59.9 & 30.4 / 57.7   \\
            + NER correction & 38.1 / 64.6 & 33.4 / 62.6 & 35.8 / 59.9 & 31.6 / 57.6  \\ \midrule
            \textbf{W.+Rule-error DST} & 33.8 / 63.7 & 28.2 / 61.7 & 36.7 / 70.4 & 30.0 / \textbf{67.2} \\
            + normalization & 36.3 / 63.1 & 31.6 / 61.2 & 40.3 / 69.4 & 34.5 / 66.6 \\
            + NER correction & 37.2 / 62.9 & 32.6 / 61.0 & \textbf{40.9} / \textbf{69.4} & \textbf{35.5} / 66.4 \\ \midrule
            \textbf{W.+TTS-ASR DST} & 33.9 / 61.6 	& 29.3 / 61.0	& 31.0 / 58.5 	& 27.8 / 56.8 \\
            + normalization & 38.1 / 63.7 & 34.4 / 63.2 & 36.0 / 60.2 & 32.0 / 58.6 \\
            + NER correction & \textbf{38.8} / 63.8 & \textbf{34.7} / \textbf{63.3} & 36.4 / 60.2 &  32.6 / 58.5  \\ \bottomrule
        \end{tabular}
    }
    \caption{Main JGA / STA results for spoken DST. Note that Text Oracle shows the upper-bound performance of the model and W. stands for Whisper.}
    \label{tab:main_results}
    \end{table}
    
    There is a large gap in JGA between the oracle and ASR results, but as we observe a small difference in terms of STA (respectively 65.7\% and 63.2\%, of which around 40\% are omissions), we can already narrow down the issue to value prediction rather than slot identification. We note that both improvements to the transcriptions and to the state tracking contribute to boost performances, especially for human speech, but do not completely recover the difference with the oracle. 
    
    We further focus on the \textbf{Dev-Human} set since it is the targeted setting and, unless mentioned otherwise, we use \textbf{W.+TTS-ASR DST} with normalization and NER correction.
    
    \subsection{What is different in spoken DST?}
    
\begin{figure}[htb]
    \centering
    \includegraphics[width=1\linewidth]{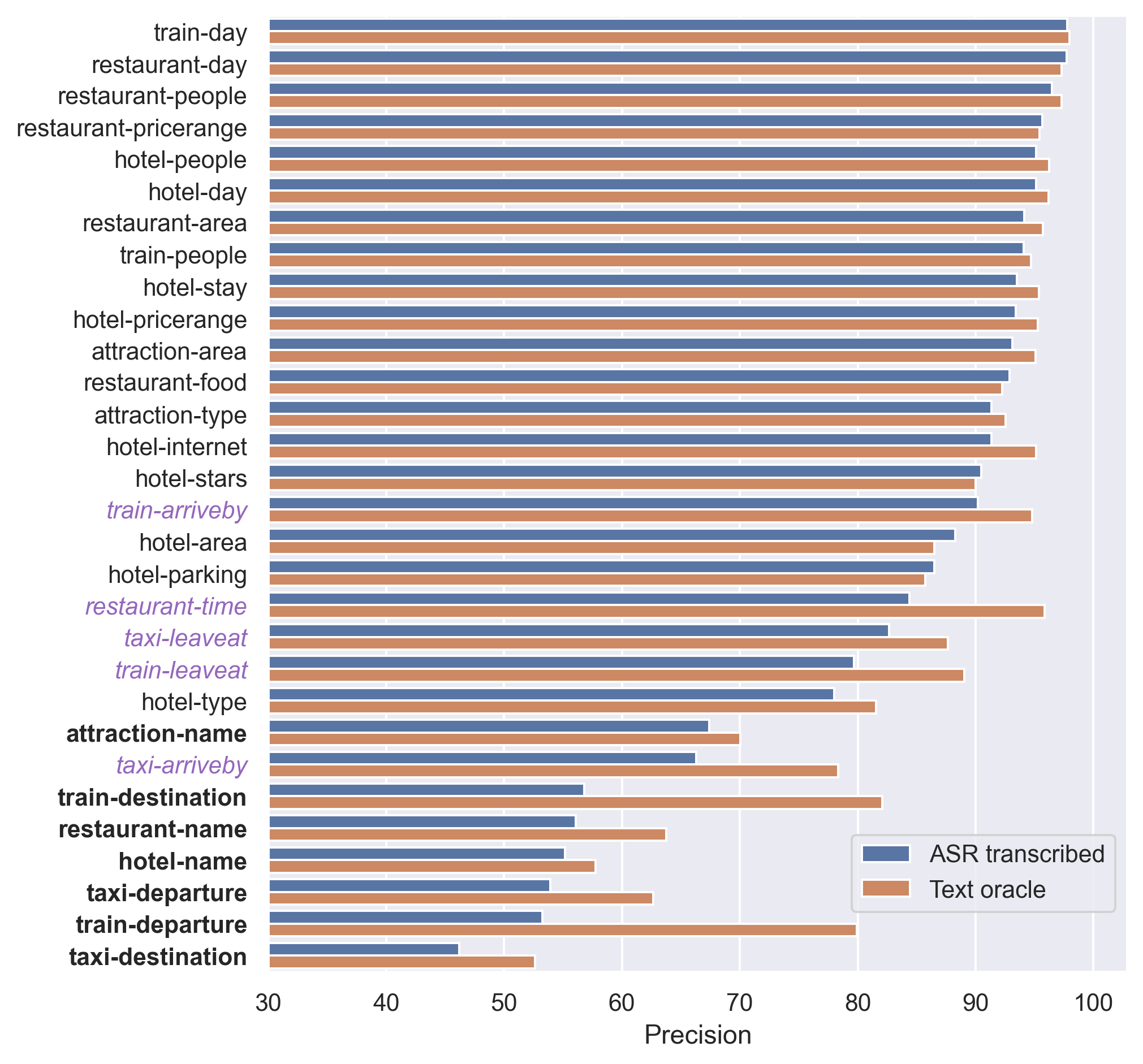}
    \caption{SP comparison between DST predictions on the text oracle and the Whisper transcriptions of the human dev set. Non-categorical slots are shown in bold and time slots in purple.}
    \label{fig:slot_acc}
    \end{figure}

    In order to explain the gap observed between the performance on the ground-truth transcriptions compared to the predicted transcriptions we present each Slot-type's value Precision in Fig. \ref{fig:slot_acc}. 
    
    Non-categorical slots and time slots explain most of this performance gap with locations (\textit{i.e.} values of slots which concern \texttt{departure} and \texttt{destination} of train or taxi trips) being particularly challenging. We observe around an 18\% macro-averaged SP decrease for non-categorical slots, a 10\% decrease for time slots, and a 1\% decrease for categorical slots between DST with ground-truth transcriptions and with ASR transcriptions.
    
    Considering that the main leverage lies in improving predictions for non-categorical slots, the remainder of our analysis focuses on said slots.  
    
    \subsection{Where do these errors come from?}
    
    
    To get a better sense of what kind of errors affects the system overall, we categorize errors for non-categorical slots predictions based on whether the predicted value in the dialogue state matches the reference value, and if the reference value can be found in the dialogue history.
    A large proportion of values are incorrect in both the context and the dialogue state (\textbf{DS no match, Context no match}) and some values are not correctly identified by the DST module while they can be found in the context (\textbf{DS no match, Context match}). We can also note that the generative nature of the DST module enables it to correct some mistranscribed values (\textbf{DS match, Context no match}).
    

    
\begin{figure}[htb!]
    \centering
    \includegraphics[width=1\linewidth]{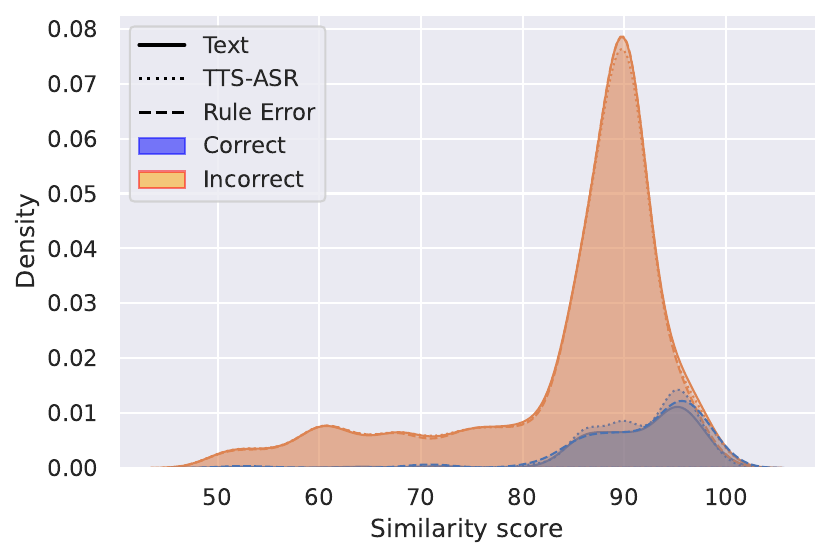}
    \caption{Distribution of the similarity scores between ASR mistranscribed values in the dialogue history and the corresponding gold values whether the DST corrected the value.}
    \label{fig:similarity}
    \end{figure}

    If we focus on the case \textbf{DS match, Context no match}, one might wonder how far away the best matching span is. 
    We compute the transcribed value similarity by computing the Levenshtein distance between the ground truth $n$-gram slot value and all $n$-grams from the ASR transcriptions. The final similarity score is the score of the closest $n$-gram.
    Fig. \ref{fig:similarity} illustrates that only values which are reasonably well transcribed are rightfully corrected by the DST (blue density shifted on the right). However the DST is far from correcting all such slot values (orange density higher than blue one). 
    
     By further manually reviewing samples of corrections made by the DST model (\textit{i.e.} \textbf{DS match, Context no match}), we find that these corrections occur in two broad categories: (i) actual corrections made by the language model (\textit{e.g.} "huntington" corrected as "huntingdon", an English town), and (ii) correction due to normalized spelling in the training data (\textit{e.g.} apostrophes ignored, British English spelling, systematic misspelling of "portugese"). Additionally, we review failures of the DST model (\textit{i.e.} \textbf{DS no match, Context match}) and find that most of these errors are due to hallucinations, the main drawback of generative models (\textit{e.g.} "boney to boyd" predicted as "boisey to boise"), or confusions in the context (\textit{e.g.} inversion of departure and destination, user repair).

    \section{Towards potential solutions}
    
    \subsection{DST correction capacity}
    Based on Fig. \ref{fig:similarity}, an improvement avenue consists in improving the correction capacity of the DST module on the transcriptions close to the ground-truth values \textit{i.e.} increasing the blue density for similarity scores over 80\%. Both the Rule Error and TTS-ASR training regimes for DST slightly improves this correction capacity. Therefore investigating the mechanisms behind such corrections might help us improve spoken dialogue systems performance. A first observation is that only values with high similarity with their target are corrected.
    
    \subsection{Importance of context}
    End-to-End approaches might reduce cascading errors thanks to joint-optimization, however such systems are challenging to design since they require careful context propagation given that audio inputs are larger than textual inputs. To highlight this point we assess the importance of context by using the ground-truth transcriptions of the user turns in all but the current turn and comparing that with all user turns transcribed by the ASR. The results in Table \ref{tab:context} show that using the ground-truth transcriptions in the previous context brings a +6 JGA increase.
    
    
    \begin{table}[hbt!]
    \resizebox{\linewidth}{!}{%
    \begin{tabular}{lc|c}
    \toprule
            & All turns as ASR hyp. & Text oracle for turns $0:t-1$ \\ \midrule
    JGA & 34.0    & 40.1       \\ \bottomrule
    \end{tabular}%
    }
    \caption{Analysis of the importance of the dialogue history.}
    \label{tab:context}
    \end{table}
     
    
    \section{Conclusion}
    Since DSTC2 \cite{henderson-etal-2014-second}, ASR systems have greatly improved, however, this has not led to a corresponding improvement for spoken DST. Through our error analysis, we have underlined that the main leverage point to bridge the gap between text and spoken DST is non-categorical slots' values. For such slots values, we have identified the need for both reducing ASR errors distance to target and increasing DST corrections. While our error injection data augmentation strategy slightly increases this correction capacity, further work in this direction is needed.
    
    The issue of open-set values should also be investigated in a complete dialogue system scenario which includes a realistic database of potential entities of interest to assess whether these errors can be corrected with thanks to the database.
    
    While for computational reasons our study focuses on Whisper-Medium and T5-Small models, we achieved a 3 to 4 points increase in JGA with larger models \cite{olisia}. The way our conclusions scale with model size remains to be studied in future work.

    \bibliographystyle{IEEEbib}
    \bibliography{bib}

\end{document}